\documentclass[sigconf,table]{acmart}

\AtBeginDocument{
  }

\copyrightyear{2024} 
\acmYear{2024} 
\setcopyright{rightsretained} 
\acmConference[ICMI '24]{INTERNATIONAL CONFERENCE ON MULTIMODAL INTERACTION}{November 4--8, 2024}{San Jose, Costa Rica}
\acmBooktitle{INTERNATIONAL CONFERENCE ON MULTIMODAL INTERACTION (ICMI '24), November 4--8, 2024, San Jose, Costa Rica}\acmDOI{10.1145/3678957.3685757}
\acmISBN{979-8-4007-0462-8/24/11}

\usepackage{colortbl}

\begin{document}

\title[Multilingual Dyadic Interaction Corpus NoXi+J]{ Multilingual Dyadic Interaction Corpus NoXi+J: \texorpdfstring{\\}{} Toward Understanding Asian-European Non-verbal Cultural Characteristics and their Influences on Engagement}

\author{Marius Funk}
\affiliation{
  \institution{University of Augsburg}
  \city{Augsburg}
  \country{Germany}
}
\email{marius.funk@uni-a.de}
\orcid{0009-0008-3064-599X}

\author{Shogo Okada}
\affiliation{
  \institution{Japan Advanced Institute of Science and Technology}
  \city{Nomi}
  \country{Japan}
}
\email{okada-s@jaist.ac.jp}
\orcid{0000-0002-9260-0403}

\author{Elisabeth André}
\affiliation{
  \institution{University of Augsburg}
  \city{Augsburg}
  \country{Germany}
}
\email{andre@informatik.uni-augsburg.de}
\orcid{0000-0002-2367-162X}

\renewcommand{\shortauthors}{Funk et al.}

\begin{abstract}
Non-verbal behavior is a central challenge in understanding the dynamics of a conversation and the affective states between interlocutors arising from the interaction. Although psychological research has demonstrated that non-verbal behaviors vary across cultures, limited computational analysis has been conducted to clarify these differences and assess their impact on engagement recognition.
To gain a greater understanding of engagement and non-verbal behaviors among a wide range of cultures and language spheres, in this study we conduct a multilingual computational analysis of non-verbal features and investigate their role in engagement and engagement prediction. To achieve this goal, we first expanded the NoXi dataset, which contains interaction data from participants living in France, Germany, and the United Kingdom, by collecting session data of dyadic conversations in Japanese and Chinese, resulting in the enhanced dataset NoXi+J. Next, we extracted multimodal non-verbal features, including speech acoustics, facial expressions, backchanneling and gestures, via various pattern recognition techniques and algorithms. Then, we conducted a statistical analysis of listening behaviors and backchannel patterns to identify culturally dependent and independent features in each language and common features among multiple languages. These features were also correlated with the engagement shown by the interlocutors. Finally, we analyzed the influence of cultural differences in the input features of LSTM models trained to predict engagement for five language datasets. A SHAP analysis combined with transfer learning confirmed a considerable correlation between the importance of input features for a language set and the significant cultural characteristics analyzed.
\end{abstract}



\received{May 10th, 2024}
\received[accepted]{July 18th, 2024}
\received[revised]{August 16th, 2024}

\maketitle
\section{Introduction}
Considering cultural differences in non-verbal behavior is essential for seamless conversations in different languages. This problem has been discussed as far back as Edward T. Hall in 1959, who stressed the importance of \textit{"the non-verbal language which exists in every
country of the world and among the various groups within
each country"} \cite{Hall1959TheSL} to understand the ways of interaction between different cultures.

The field of non-verbal behavior and backchannels in culture-dependent human-human interaction has since been extensively studied, whereas non-verbal behavior in human-computer interaction has focused predominantly on single-culture human-computer interaction \cite{MultiMediate,ArtiranDetectionHeadNods,Gaze1,japModel1}. Even though it is acknowledged as important, cultural characteristics have not been a focus in Social Signal Processing \cite{Vinciarelli2012}.

In this paper, we present a computational analysis of the cultural characteristics of multimodal non-verbal features and the effects these differences have on engagement and its prediction. For this purpose, we introduce a new multilingual multimodal interaction dataset, \textbf{NoXi-J}, which enhances the existing \textbf{NO}vice e\textbf{X}pert \textbf{I}nteraction database \textbf{NoXi} \cite{NoXi} by recording sessions in \textbf{J}apanese and Chinese, thereby creating an enriched dataset referred to as \textbf{NoXi+J}. We also extract and analyze the non-verbal features and vocal backchannels of all predominant languages (German, English, French, Japanese, and Chinese) using various machine learning models and pattern recognition techniques. We study the individual multimodal non-verbal features and investigate the differences between language sets and their correlation with engagement.
Finally, we highlight the importance of culture-sensitive approaches with machine learning engagement prediction models. We evaluate the performance of six models trained on different language speaker subsets of the NoXi+J dataset, test the model performance on other language speaker subsets and compare the performance of the model depending on the relevant non-verbal and backchanneling features.

To the best of our knowledge, there is no other comprehensive data-based analysis of the cultural differences in backchanneling and non-verbal communication and their influences on engagement in a recorded multimodal database. The addition of the Japanese and Chinese language recordings also makes it the largest openly available multicultural multimodal corpus of dyadic interaction of which we are aware.

In the following pages, we briefly describe \textbf{NoXi+J}, its design process, recording system, and data; we focus on the newly collected data, and outline the manually created affective annotations. Next, we computationally analyze the data, focusing on non-verbal features, backchanneling, and speaking state and their influences on engagement. Finally, we describe a set of engagement prediction models trained on various language subsets of the complete dataset, how the difference in their performance correlated with the results of the analysis and how transfer learning affects their performance.

\section{Scientific Background}

\subsection{Non-verbal communication}

Non-verbal communication in the context of conversations involves gestures, postures, touch, facial expressions, gaze, and vocal behavior beyond the meaning of words \cite{knapp1972nonverbal}. Non-verbal communication can manage the flow of a conversation and therefore the turn-taking \cite{matsumoto2012nonverbal} and influence engagement by conveying emotional states and signaling interpersonal attitudes \cite{LAFRANCE197871}.

Cultural differences in non-verbal behavior have been acknowledged for a long time \cite{hall1976beyond,Andersen1998-al}. Research often focuses on facial emotions \cite{Matsumoto1993, Ekman1994} and differences between East Asian and European facial expressions \cite{Jack2012,Matsumoto1993}. 
Another focus of non-verbal communication is head-nodding, where cultural differences, especially the prevalence of Japanese head nods, are often noted \cite{FREIERMUTH2023103629,KITA2007}.

\subsection{Backchanneling}\label{Backchanneling}

The concept of listener utterances that do not lead to turn-taking has been discussed as far back as Fries in 1952 \cite{fries1952}, whereas the term \textit{backchannel} to describe this kind of verbalisation was introduced by Yngve in 1970 \cite{Yngve70}. Backchannels often consist of utterances such as the English \textit{uh huh} and \textit{yeah} \cite{White_1989} but can also be longer phrases such as the Japanese \textit{sou desu ne} \cite{Hanzawa}. They can also include head nods \cite{MAYNARD1987} or laughter and exhaling sounds \cite{Aizuchi}. 

Studies often highlight that Japanese interlocutors use back\-channels with much higher frequency compared to English or Chinese speakers\cite{White_1989,MAYNARD1990397,Cutrone2005,KITA2007}.

\subsection{Turn-taking}

Turn-taking describes the changing of the active speaker in a conversation. Each time such a change takes place is classified as an instance of turn-taking \cite{TurnWiemann}. The role division of the active speaker and listener and the issues that arrive when overlapping talk occurs significantly impact the course of conversations \cite{SCHEGLOFF_2000}. 

Turn-taking is crucial for the management of the flow of conversations and is often indicated by non-verbal communication \cite{SKANTZE2021101178}. Pauses can also indicate turn-taking, although short pauses between speech are common without turn-taking occurring \cite{BOSCH200580}. Turn-taking and its timing have been shown to noticeably influence engagement \cite{ChaoTiming,CafaroInterruptingEngagement}.

\subsection{Engagement} \label{Engagement}

Engagement refers to the interest a person shows in an ongoing conversation or interaction.
It can be measured either continuously or at specific interaction points. It can be assessed between participants or for individual interlocutors separately. Engagement may be directed toward a human interlocutor, a system, or an artificial agent \cite{Definitionsofengagement}.  

One of the earliest definitions in the context of human-computer interaction comes from Sidner et al. \cite{SIDNER2005140}, who describes it as
\textit{"the process by which individuals in an interaction start, maintain and end their perceived connection to one another"}. Sidner et al. emphasize the role of non-verbal behavior and turn-taking as indicators of engagement. 
In the context of dyadic conversations, Poggi \cite{poggi2007mind} defines engagement as
\textit{"the value that a participant in an interaction attributes to the goal of being together with the other participant(s) and of continuing the interaction"}.

\section{Related Work}
In recent years, the analysis and prediction of non-verbal communication, turn-taking and backchannels have gained importance in interaction modeling \cite{ortega2023modeling,JainExploring, BRUSCO202024,Levinsonturn-takingmodels}. Researchers have focused on gaze and its role in recognizing intention \cite{Gaze1}, how non-verbal actions signal human preferences \cite{CandonNonverbal}, the estimation of agreement \cite{mueller:et:al:2022}, head nod detection \cite{ArtiranDetectionHeadNods} and backchannel prediction using multimodal approaches \cite{Wang2024,japModel1, MultiMediate}.

Attempts have been made to equip virtual agents and robots with culture-specific behaviors. In this context, we refer to a survey that reports on how emotions are portrayed in different cultures and explores how virtual agents and robots can simulate culture-specific emotional behaviors \cite{Andre2014}. Endrass et al. \cite{Endrass2013} developed computational models to replicate prototypical behaviors of German and Japanese cultures in virtual agents, taking into account verbal behavior, communication management, and non-verbal behavior. Meixuan et al. \cite{MeixuanLI2020} collected annotated voice responses in three languages — Chinese, English, and Japanese — with the aim of developing emotionally attuned robot models.

Many multimodal datasets focus on behavioral and emotional analysis, such as the Cardiff Conversation Dataset \cite{AubreyCardiffConversation}, which contains 30 conversations with annotation for head movement, speaker activity, and non-verbal utterances, or SEMAINE \cite{McKeownSEMAINE}, which features 150 recordings with emotional annotations. However, multicultural conversation datasets for comprehensive non-verbal analysis are rare as researchers often focus on text analysis \cite{Qiu_Zhao_Li_Lu_Peng_Gao_Zhu_2022} or present study results without making their datasets publicly available \cite{workplaceinterculturalHe}. A few examples of datasets featuring multicultural interactions are the RECOLA Dataset \cite{RingevalRECOLA}, which features collaborative and affective interactions with French, German, Italian and Portuguese participants, and
the Sentiment Analysis in the Wild (SEWA) dataset, which contains recordings of
British, German, Hungarian, Greek, Serbian, and Chinese participants \cite{Kossaifi_2021}.

Several reviews indicate the growing interest in engagement and its significance in human-computer interaction \cite{DohertyEngagement, GlasDefinitionsEngagement,Oertel2020}. Various techniques for engagement prediction \cite{bohus-horvitz-2009-learning, InoueLatent, NakanoEstimating} have been developed as part of interactive systems. 

Research on the prediction of engagement has already been conducted on the original NoXi dataset \cite{LiInter-person, DermoucheEngagement}. However, the focus of this work was on the performance in the prediction task and not on the analysis of culture-specific aspects of the dataset.

\section{Data Collection}
The original NoXi database, first introduced by Cafaro et al. \cite{NoXi}, contains recordings of dyadic conversations between two interlocutors in the roles of expert and novice. In a session ranging between seven and 31 minutes the expert talks about one or more topics they are passionate or knowledgeable about, whereas the novice listens and discusses the introduced topic with the expert. The idea was to obtain a dataset of natural interactions in an expert-novice knowledge-sharing context.

The database contains conversations primarily in German, English, and French. NoXi-J extends it by adding 48 conversations in Japanese and 18 conversations in Chinese to provide a more culturally diverse dataset.
The newly recorded sessions follow the same structure as the original sessions. We will briefly explain the design principles and the recording system used to record the new sessions. For a more detailed explanation, see the initial paper \cite{NoXi}.

\subsection{Design Principle}

\begin{figure}[t]
\centerline{\includegraphics[width=1\columnwidth]{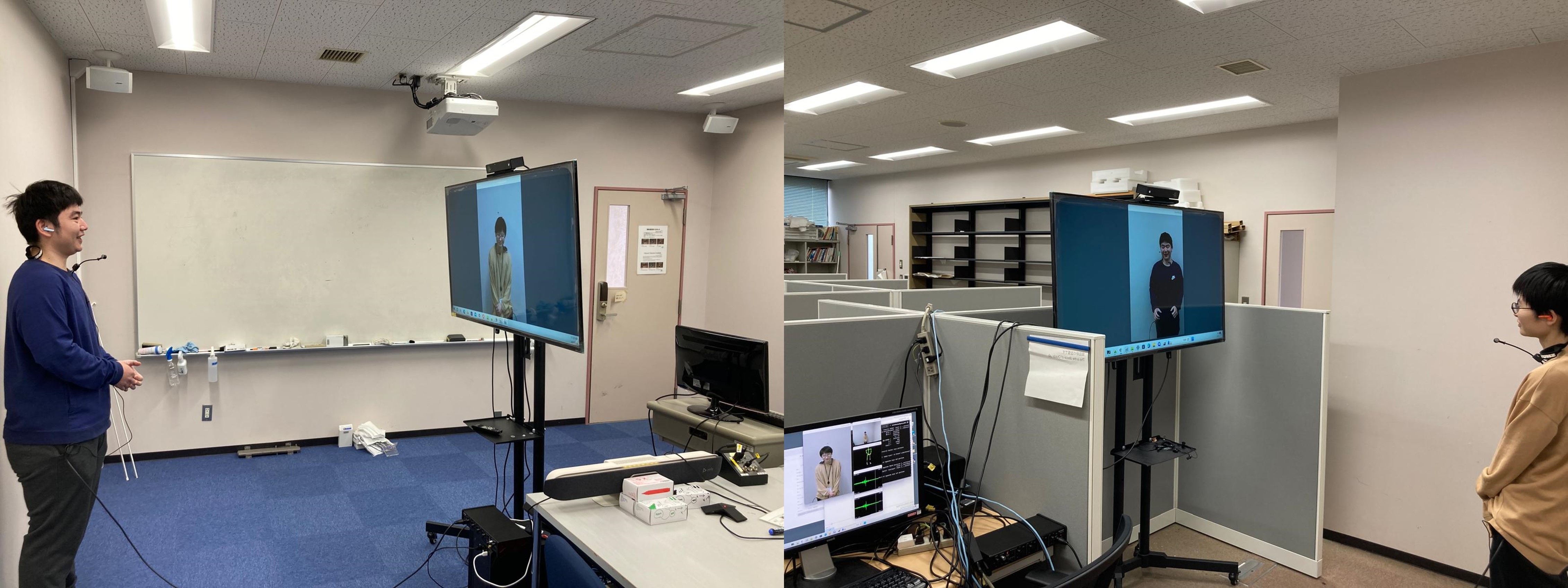}}
\caption{NoXi recording. Expert (left) and novice (right).}
\label{setup}
\end{figure}

\subsubsection{Setting} 
Screen-mediated recording was chosen for two purposes: to record a face-to-face conversation without requiring multiple cameras recording from different angles and to create a setup more similar to an interaction between a human and a virtual agent.  To ensure the capture of facial expressions, gestures, speech and full body movements (e.g. head touch), the participants were recorded in a standing position.
The setup used for NoXi and NoXi-J was nearly the same, with minor changes, such as the use of headphones instead of speakers to reduce echo in the case of NoXi-J.

\subsubsection{Interaction} The recordings consist of spontaneous interactions that are focused on knowledge transfer and information retrieval but also include planned occurrences of unexpected events (e.g. interruptions). The conversations were not interrupted after the target length of 10 minutes, leading to some interactions exceeding 30 minutes. The actual setup of the interactions for the Asian recordings can be seen in Figure \ref{setup}.

\subsubsection{Participants} In the European sessions, participants were recruited from local research facilities and their immediate social circles. For the newly recorded Asian sessions, participants were also recruited from local research facilities and, for many Japanese sessions, through an employment agency. This approach provided a wide variety of relationships between expert-novice dyads, ranging from zero-acquaintance situations \cite{Ambady1995} to spouses.

\subsubsection{Unexpected Events} One of the goals was to obtain occurrences of unexpected events. In addition to events such as spontaneous debates during the session, we artificially injected unexpected events during the recording sessions. These events were one of two types of interruptions. Approximately five minutes after the start of a session we either called the novice on their mobile phone (i.e., CALL-IN) or physically entered the recording room to adjust the microphone or ask them to hand over something (i.e., WALK-IN). The novice was informed about the possibility of one of these events occurring prior to the session, in contrast to the expert, who was intended to be surprised and annoyed by the interruption.

\subsubsection{Recording Protocol}
The recording protocol had slight differences between the European and Asian parts of the corpus. For the European recordings, the participants were received and instructed in different rooms, whereas for the Asian recordings, the initial explanation was given in a shared room before the participants were split into different rooms. 
We then primed the novice about the functional interruption, set up their microphone, and indicated where the participants had to stand (also indicated by a marker on the floor or wall). The session was monitored in a separate room.  
After the conversation concluded naturally, the participants were given questionnaires (see Section \ref{collecteddata}), informed about their compensation, and debriefed.

Both participants gave their informed consent before the start of the recording. They consented to the use of the recorded data for scientific research and noncommercial applications. The participants had three choices regarding the usage of their data. All participants agreed to (1) the use of their data within the PANORAMA project consortium.
Additionally, most participants agreed to (2) the usage of the data in academic conferences, publications and/or as part of teaching material and to (3) the usage of the data for academic and non-commercial applications to third-party users internationally, provided that those parties upholding the same ethical standards as the PANORAMA Project.

\subsection{Recording System}

The data were recorded using Microsoft Kinect 2 devices for full HD video streams and ambient noise capture. Furthermore, low-noise recordings of voices were obtained using dynamic head-set microphones (Shure WH20XLR connected via a TASCAM US-322). The Kinect devices were placed over 55" flat screens. Both were connected to PCs (i7, 16GB-32GB of RAM). Each room's system captured and stored the recorded footage and signal streams locally. A third PC observed the interaction from another room. All PCs were connected over LAN. 

To sync the recordings, a two-step synchronization was used. Once all the sensors were connected and the setup was completed, we used a network broadcast from the observer room PC to start recording in the expert's room and novice's room simultaneously. The system was implemented with the Social Signal Interpretation framework \cite{WagnerSSI}. For more details regarding the setup and the frameworks used, we refer to the introductory paper of NoXi \cite{NoXi}.

\subsection{Collected Data} \label{collecteddata}

\begin{table}[t]

\begin{center}
\begin{tabular}{|c|c|c|c|c|c|}
\hline
\textbf{Lang.}&\textbf{Ses.}&\textbf{Part.}&\textbf{Avg. Dur.}&\textbf{Std. Dur.}&\textbf{Tot. Dur.} \\
\hline
DE& 19& 29 (5/24) & 17:56 & 05:56 &  05:38 \\
FR& 21& 32 (8/20) & 20:12 & 06:35 &  07:20 \\
EN& 32& 26 (11/14) & 16:49 & 05:55 &  08:35 \\
JP& 48& 48 (18/30) & 14:30 & 04:06 &  11:36 \\
ZH& 18& 18 (10/8) & 15:16 & 03:22 &  04:35 \\
Other& 12& 18 (5/5) & 17:37 & 07:09 &  03:28 \\

\hline
Total & 150 & 153 (54/99) & 16:36 & 05:44 & 41:11 \\
\hline

\end{tabular}

\vspace{4mm} 
\caption{Overview of all the recorded NoXi sessions. From left to right: Language of the recording, number of recording sessions, number of participants (female/male), average and standard deviation of the recording duration (mm:ss), and the total duration (hh:mm). Some participants were used in sessions of multiple languages.}
\label{sessionOverview}
\end{center}
\end{table}

\begin{figure}[t]
\centerline{\includegraphics[width=1\columnwidth]{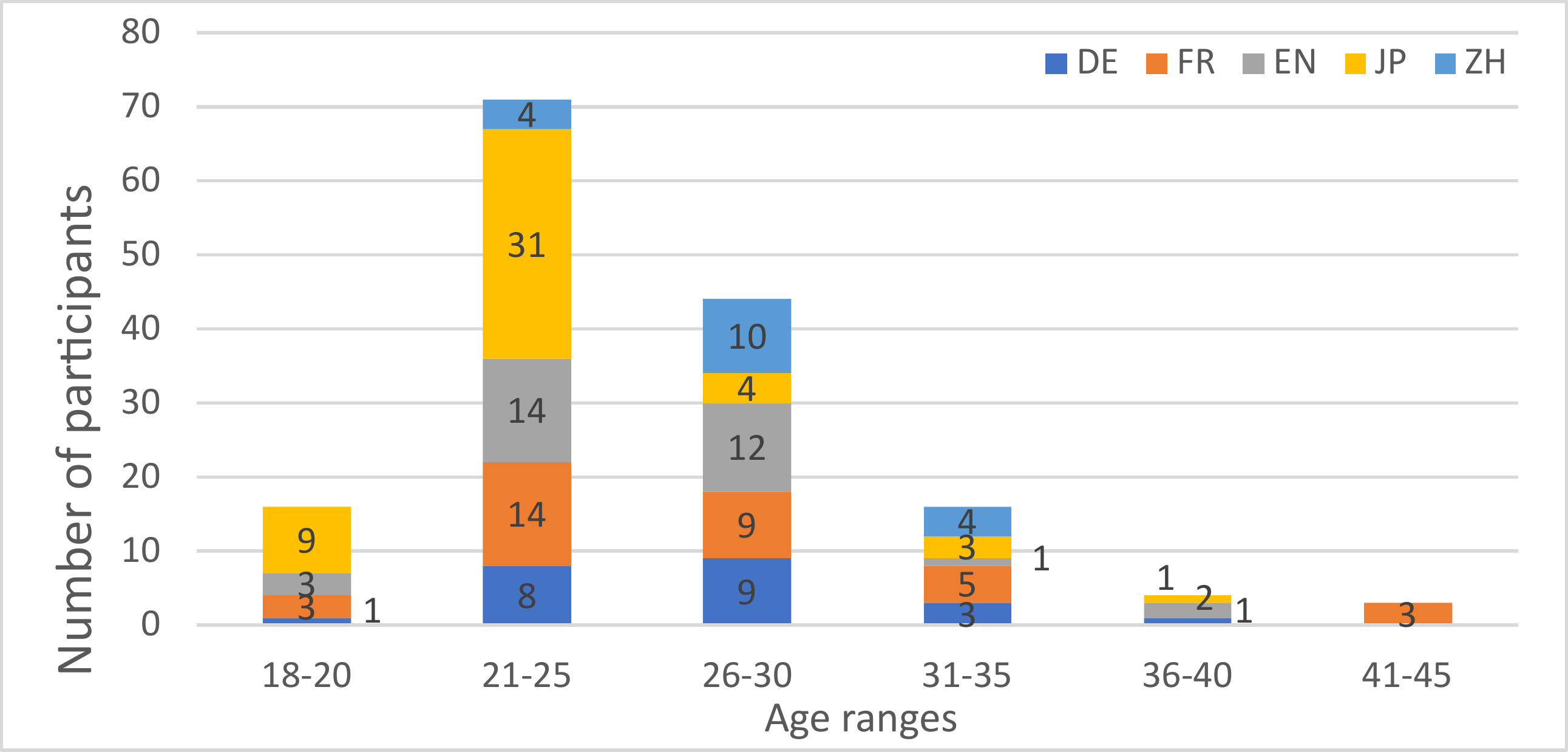}}
\caption{Age distribution of the speakers of the 5 primary recorded languages: German (DE), French (FR), English (EN), Japanese (JP) and Chinese (ZH). \cite{NoXi}}
\label{agedist}
\end{figure}

The experiment was administered in four countries, with NoXi being conducted in France, Germany, and the UK, and NoXi-J being conducted in Japan. In addition to the Japanese sessions, we decided to increase the diversity and to supplement the dataset with Chinese sessions, as many native Chinese speakers were available at the recording location. Besides the five primary languages, a smaller number of recordings of four other languages (Spanish, Indonesian, Italian and Arabic) was also collected. A summary of the recorded sessions divided by primary language can be found in Table \ref{sessionOverview}. 
In addition to session details, we recorded demographic information about the participants including their age and gender as well as their self-assigned cultural identity. A breakdown of the five primary languages and their age distributions can be seen in Figure \ref{agedist}.
We collected the discussed topics, proficiency of the language spoken for both participants, and the social relationship level between the two participants (e.g. zero acquaintance, friends). All participants provided a self-assessment of their personality on the basis of the Big 5 model \cite{Goldberg1990} by using descriptions for Saucier's Mini-Markers set of adjectives (consisting of 40 adjectives) \cite{Saucier1994}.

Anonymized data are available for the NoXi+J dataset upon request from the authors at the e-mail address noxi+j@hcai.eu.
\subsection{Annotations}
Over 40 annotators from 4 countries (Germany, France, the UK, and Japan) were involved in the annotation of the NoXi+J database. Annotations were made and managed using the freely available annotation tool NOVA\footnote{https://github.com/hcmlab/nova}.

In this paper, we focus exclusively on the manually created engagement annotations (see Section \ref{Engagement}). Similar to the original NoXi corpus, engagement in the NoXi-J dataset has been annotated by three or more individuals. However, this excludes the complete Chinese language set and many Japanese language sessions, which were only recently recorded, and contain fewer annotations. To minimize annotator bias, the average of all annotations for each session and role (expert or novice) is used for further analysis. The engagement annotations assign values between 0 and 1 to every frame of the dataset for the perceived engagement at that moment. 

Some differences may arise from annotator bias \cite{Hovy2021}. 
This is unavoidable as establishing a definitive ground truth is not possible, not even with self-reporting \cite{questforGroundtruth}. 
To determine the extent of annotator bias, we calculated the intercoder reliability \cite{IntercoderO’Connor}. Using the ICCk3 value of the intraclass correlation coefficient, we calculated an overall intercoder agreement of 0.63. The Mean Absolute Error between annotators was between 0.14 and 0.15 for the engagement scores of 0-1 for all annotated languages. 

\section{Data Analysis} \label{DataAnalysis}

\begin{table}[t]
\caption{List of all used features for novice and expert.}
\resizebox{\columnwidth}{!}{
\begin{tabular}{|c|c|l|}
\hline
\textbf{Feature}&\textbf{Dim}&\textbf{Explanation}\\
\hline
Engagement& 1 & Continuous annotation of engagement\\
\hline
Body Properties& 20 & 20 different body properties such 

as \textit{Arms open}, \textit{Energy Head}, etc. \\
\hline
Action Units& 17 & AU26 (Jaw Open) \\
&  & AU18 (Lip Pucker) \\
&  & AU30 (Jaw Slide) \\
&  & AU20 (Lip Stretcher) (Left,Right) \\
&  & AU12 (Lip Corner Puller) (Left,Right) \\
&  & AU15 (Lip Corner Depressor) (Left,Right) \\
&  & AU16 (Lower Lip Depressor) (Left,Right) \\
&  & AU13 (Cheek Puff) (Left,Right) \\
&  & AU43 (Eye Closed) (Left,Right) \\
&  & AU4 (Eyebrow Lowerer) (Left,Right) \\
\hline
Fluidity& 1 & Fluidity of body movements\\
\hline
Head Rotations& 3 & Pitch, Yaw, and Roll of the head\\
\hline
Spatial Extent& 1 & Usage of space by movement\\
\hline
Overall Activation& 1 & Overall movement\\
\hline
Voice Activation& 1 & Instances of human speech\\
\hline
Transcription& 1 & Transcription of speech (Only for reference)\\
\hline
Head nod& 1 & Instances of intense vertical head movement \\

\hline
\end{tabular}
\label{usedFeaturesTable}
}
\end{table}

\begin{figure}[htbp]
\centerline{\includegraphics[width=1\columnwidth]{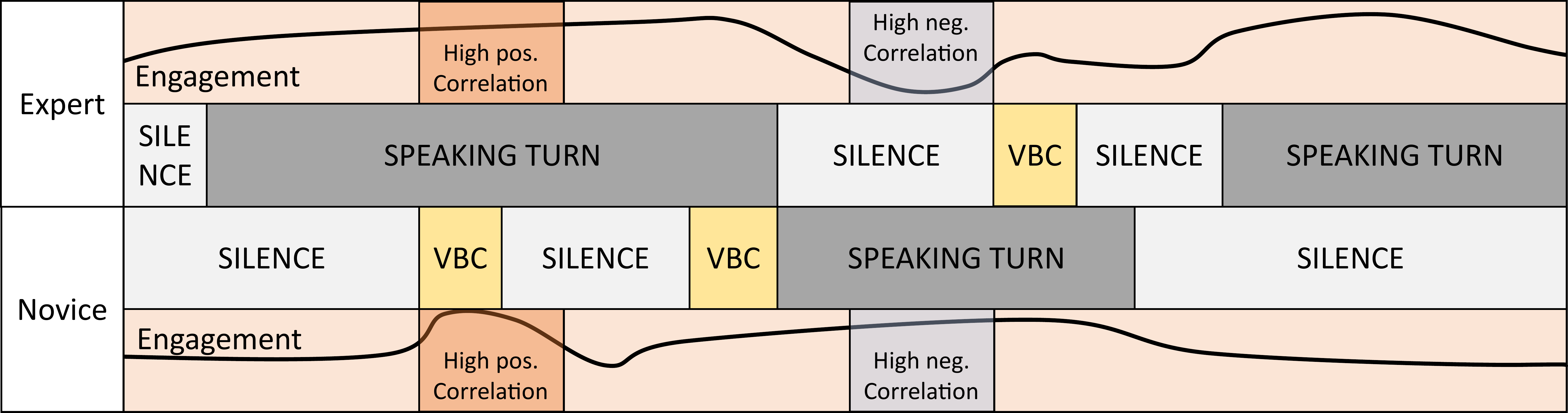}}
\caption{Schematic conversation depicting turn-taking, the division of the data by speaking state, engagement and instances of high positive and high negative engagement correlation. \textbf{VBC} describes vocal backchanneling instances.}
\label{datadivision}
\end{figure}

\subsection{Features}

\subsubsection{Features of the NoXi corpus}

For the following analysis of intercultural differences of non-verbal features, engagement, and their mutual dependencies, we decided to use a total of 94 features (see Table \ref{usedFeaturesTable}). We focus on the novice's non-verbal 
characteristics, their behavior, and the impact on engagement. However, we also analyzed expert features in relation to the novice's engagement. Engagement has not yet been annotated for the Chinese language session. Therefore only the evaluation of the feature differences will consider these recordings.

\subsubsection{Feature extractions}

In addition to the immediate output stream of Kinect such as video and joint position data
, we extracted body properties indicative of the expression of emotion \cite{BodilyexpressionWallbott} such as \textit{Head Touch}, \textit{Arms crossed}, \textit{Energy Head}, \textit{Fluidity}, \textit{Spatial Extent}, \textit{Energy Hands} and \textit{Overall Activation} of the body using NOVA's integrated extraction tools. 
For the computation of gestural expressivity features (fluidity, overall activation, spatial extent, and energy) we refer to the appendix of an earlier paper describing the NOVA annotation tool \cite{baur:et:al:2015}. We also used an external Nova Server\footnote{https://github.com/hcmlab/nova-server} with integrated Pyannote\footnote{https://github.com/pyannote/pyannote-audio} for Voice Activity Detection (VAD) and whisperX\footnote{https://github.com/m-bain/whisperX} \cite{bain2023whisperx} for the transcription of speech.

\subsubsection{Computed features}\label{compfeat}

Additional features for the analysis of these data were computed. We used the extracted voice activation data to determine turns and distinguish active speech from vocal backchanneling. We attributed the turn to the first speaker, who holds it until both interlocutors either become silent (i.e. no voice activation is determined) for two seconds (50 frames) or the speaker becomes silent after the interlocutor has spoken for more than two seconds. In the first case, no one holds the turn, and we move on to the next speaker. In the second case, turn-taking takes place. All voice activation instances in between are classified as vocal backchanneling (VBC) (see Figure \ref{datadivision}). The idea for this division was influenced by Bosch \cite{BOSCH200580} and his discussion of overlap and turn-taking.

Head nods were identified by using the pitch, yaw, and roll angles of the head position extracted from the \textit{head.stream} files. Rapid switches of over 2.5 degrees between up and down movement without extensive other movements were classified as head nods. Changes in the threshold led to different absolute numbers, while the distribution between cultures remained similar.

\subsection{Intercultural data comparison}
  
\subsubsection{Initial cultural comparison}
Our first focus was on the general assessment of the recorded features. To reduce outliers and make the data more comparable, we calculated the standard score (z-score) for every data point. We then performed an initial Analysis of Variance (ANOVA) \cite{STHLE1989259} for every feature of the session averages between languages. 
The results show an average F-value of $\sim$43.000 with 
a minimum F-value of 15.5, revealing severe differences in the features between datasets.
Table \ref{anovaTable} shows clear similarities for the inner-groups European language (NoXi) and Asian language (NoXi-J) in a subsequent pairwise Tukey-Kramer test \cite{Tukey1949,hayter1984proof}.

\begin{table}[t]
\caption{Sum of the absolute values of all the Tukey-Kramer test averages between the five languages.}
\resizebox{\columnwidth}{!}{
\begin{tabular}{|c|c|c|c|c|c|}
\hline
&\textbf{German (DE)}&\textbf{French (FR)}&\textbf{English (EN)}&\textbf{Japanese (JP)}&\textbf{Chinese (ZH)}\\
\hline
DE& \cellcolor{gray}& 16.8& 18.5 & 25.3& 28.2\\
\hline
FR&\cellcolor{gray}&\cellcolor{gray}& 14.4 & 22.4& 21.7\\
\hline
EN&\cellcolor{gray}&\cellcolor{gray}&\cellcolor{gray}&23.5&24.0\\
\hline
JP&\cellcolor{gray} & \cellcolor{gray} &\cellcolor{gray}&\cellcolor{gray}& 16.1\\
\hline
\end{tabular}
\label{anovaTable}
}
\end{table}

\subsubsection{Cultural differences between feature averages}

We found many noteworthy feature differences between cultures. The Chinese participants activated AU12, which is commonly associated with a smile, the least, followed closely by the English-speaking participants. In contrast, the Japanese smiled the most. German novices held their arms in open poses observably longer than any other group of participants. In contrast, Japanese and Chinese interlocutors adopted an arms-crossed pose for noticeably longer periods, specifically during backchanneling, while otherwise displaying behaviors similar to that of the European participants.

\subsection{Engagement in intercultural data comparison}

 Next, we analysed engagement in the annotated sessions. Overall we found that the Japanese sessions exhibit a noticeably higher average level of annotated engagement compared to the European sessions (see Figure \ref{engavg}). As no Chinese language session engagement annotations have yet been created, we discuss only German, English, French and Japanese language data from this point on.

\subsubsection{Engagement correlation with recorded and extracted features}

\begin{figure}[t]
\centerline{\includegraphics[width=1\columnwidth]{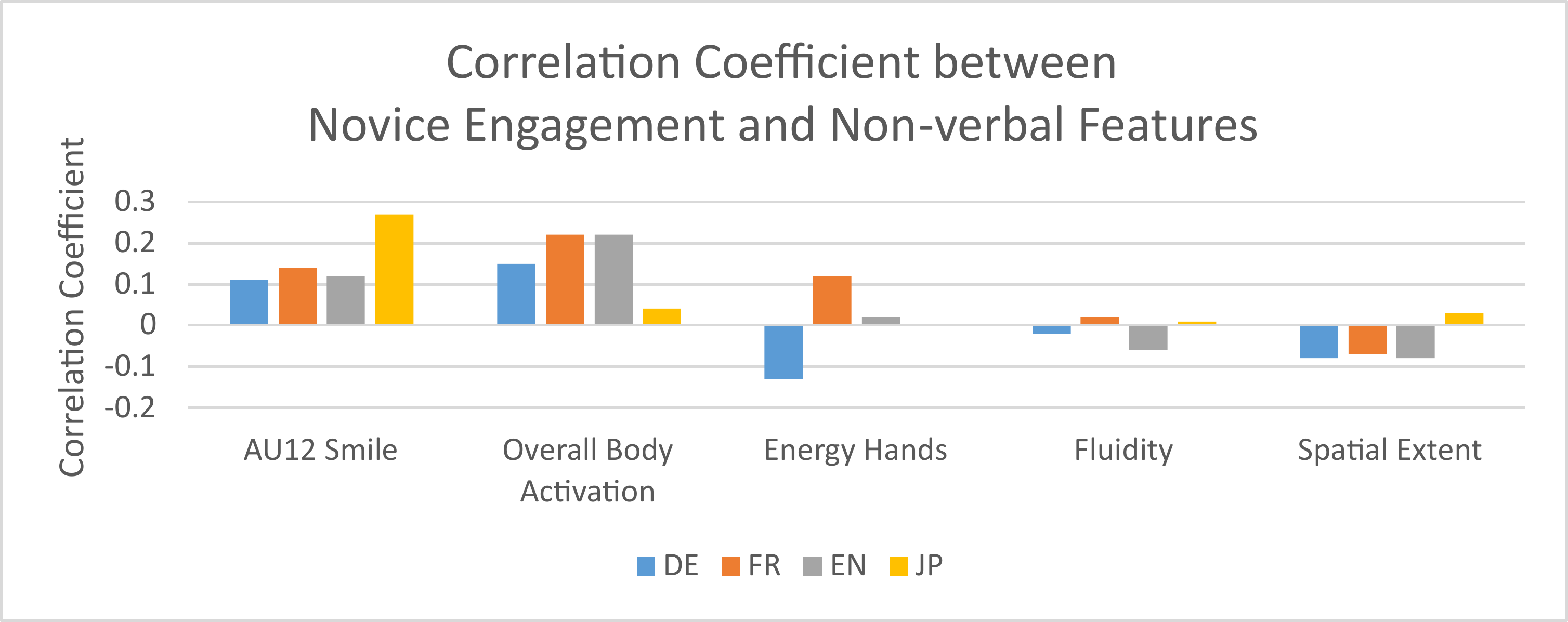}}
\caption{Correlations between annotated novice engagement and a selection of relevant features.}
\label{engfeatcorr}
\end{figure}

 We began by examining the relationship between input features and engagement within the inter-lingual dataset. Specifically, we calculated the Pearson correlation coefficient \textit{r}  between features and novice engagement as our primary metric. We found significant differences in the correlation of features and engagement between the NoXi and NoXi-J recordings (see Figure \ref{engfeatcorr}). All correlations presented in this section are statistically significant with p-values of less than 0.001. \textit{N} is the total number of frame values for each feature in each language (DE=511,200, FR=666,950, EN=784,675, JP=1,043,700).

Activation of AU12 (Smile) shows the strongest correlation with engagement in the Japanese dataset (r=0.27) compared to German (r=0.11), French (r=0.14) and English (r=0.12). 
In contrast, overall activation is more strongly correlated with engagement  the French (r=0.22), German (r=0.15) and English (r=0.22) datasets, but shows almost no correlation with the Japanese engagement (r=0.04).

A notable difference between German and French engagement correlations is observed in the energy level of the hands. While the German data show a negative correlation between engagement and hand energy (r=-0.13), the French data reveal a positive correlation (r=0.12). The other languages show no noteworthy correlation in this regard.

\subsubsection{Head nods}

\begin{table}[t]
\caption{Computed features for the novice, from left to right: Session language, registered head nods (HN), HN per minute, time ratio spent listening, and time ratio spent verbal backchanneling (VBC).}
\resizebox{\columnwidth}{!}{
\begin{tabular}{|c|c|c|c|c|}
\hline
\textbf{Language}&\textbf{Head Nods}&\textbf{Head Nods per minute}&\textbf{Listening ratio}&\textbf{VBC ratio}\\
\hline
DE& 1830 & 6.5 & 90.0\% &  6.9\% \\
\hline
FR& 2380 & 6.1 & 75.3\% &  22.1\%\\
\hline
EN& 4239 & 7.9 & 79,4\% & 14.5\% \\
\hline
JP& 5851 & 8.0 & 87.5\% & 14.5\%\\
\hline
ZH& 2167 & 7.5 &  89.6\% & 8.9\%\\
\hline

\end{tabular}
\label{headnodsTab}
}
\end{table}

We calculated the frequency of head nods on the basis of length of the recorded sessions and the recognized head nod count. We found a noteworthy difference in head nod frequency between recorded data of European participants and Asian participants (see Table \ref{headnodsTab}). 

The high frequency of the session with the English-speaking participants was caused by frequent head nodding of the participants with an Asian cultural background. The utilized algorithm could not detect small head nods obscured by static noise in the facial recognition data extracted from Kinect. An investigation of the data verified that many small head nods, especially in the Japanese dataset, were not identified. 

\subsubsection{Vocal backchannels}

Table \ref{headnodsTab} also shows the calculated ratio of listening and the time spent vocally backchanneling separately. French language interlocutors spent the least amount of time listening while engaging the most in vocal back-channeling, with similar values observed in the English-speaking sessions. 
The Japanese are unique in exhibiting a very high listening ratio and a high ratio of vocal back-channeling.

\subsubsection{Mutual engagement and speaking turns} \label{Mutualengagement}

\begin{figure}[!tbp]
  \centering
  \begin{minipage}[b]{0.5023\columnwidth}
    \includegraphics[width=\columnwidth]{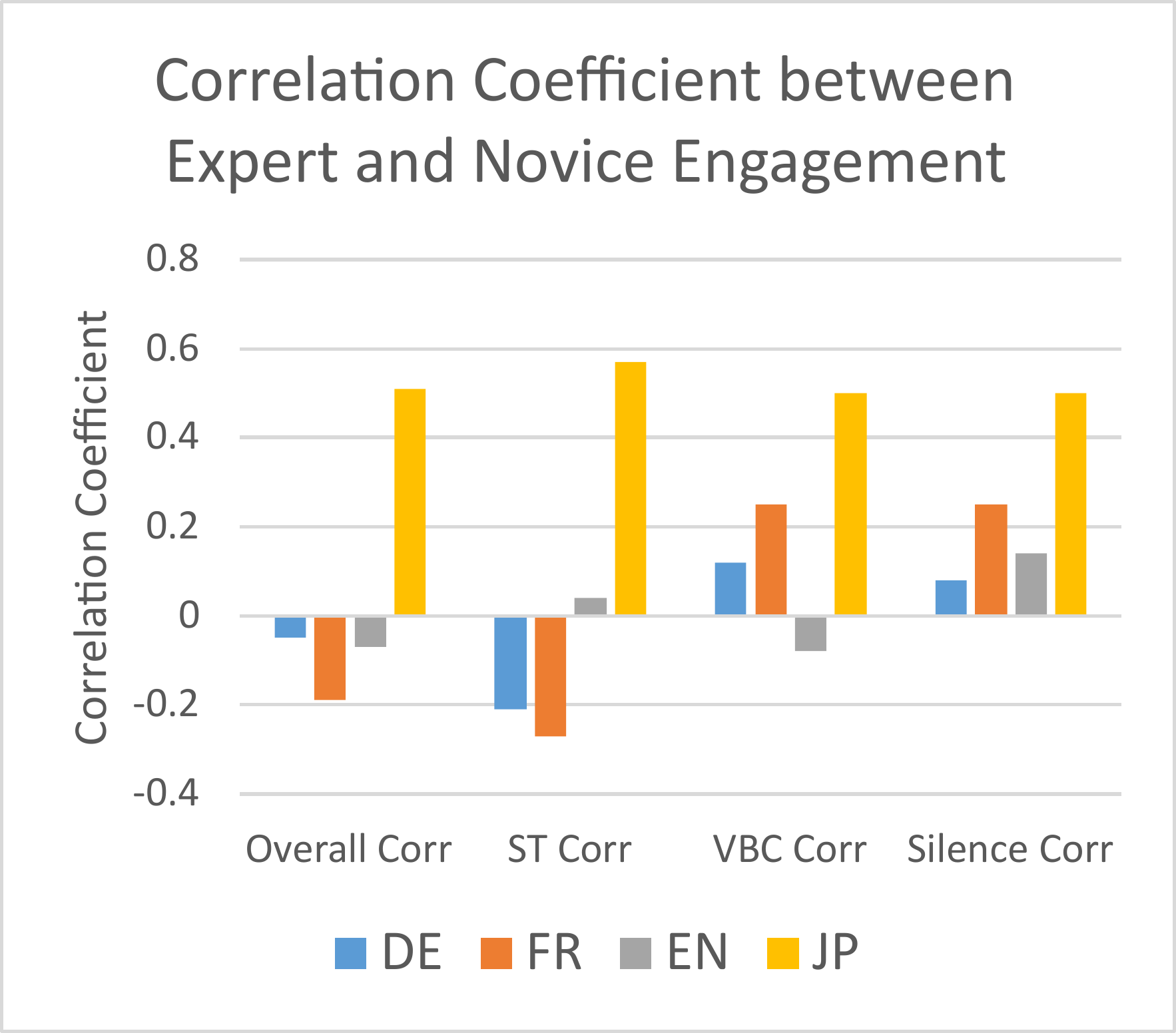}
    \caption{Correlations between expert and novice engagement.}
    \label{engcor}
  \end{minipage}
  \hfill
  \begin{minipage}[b]{0.4757\columnwidth}
    \includegraphics[width=\columnwidth]{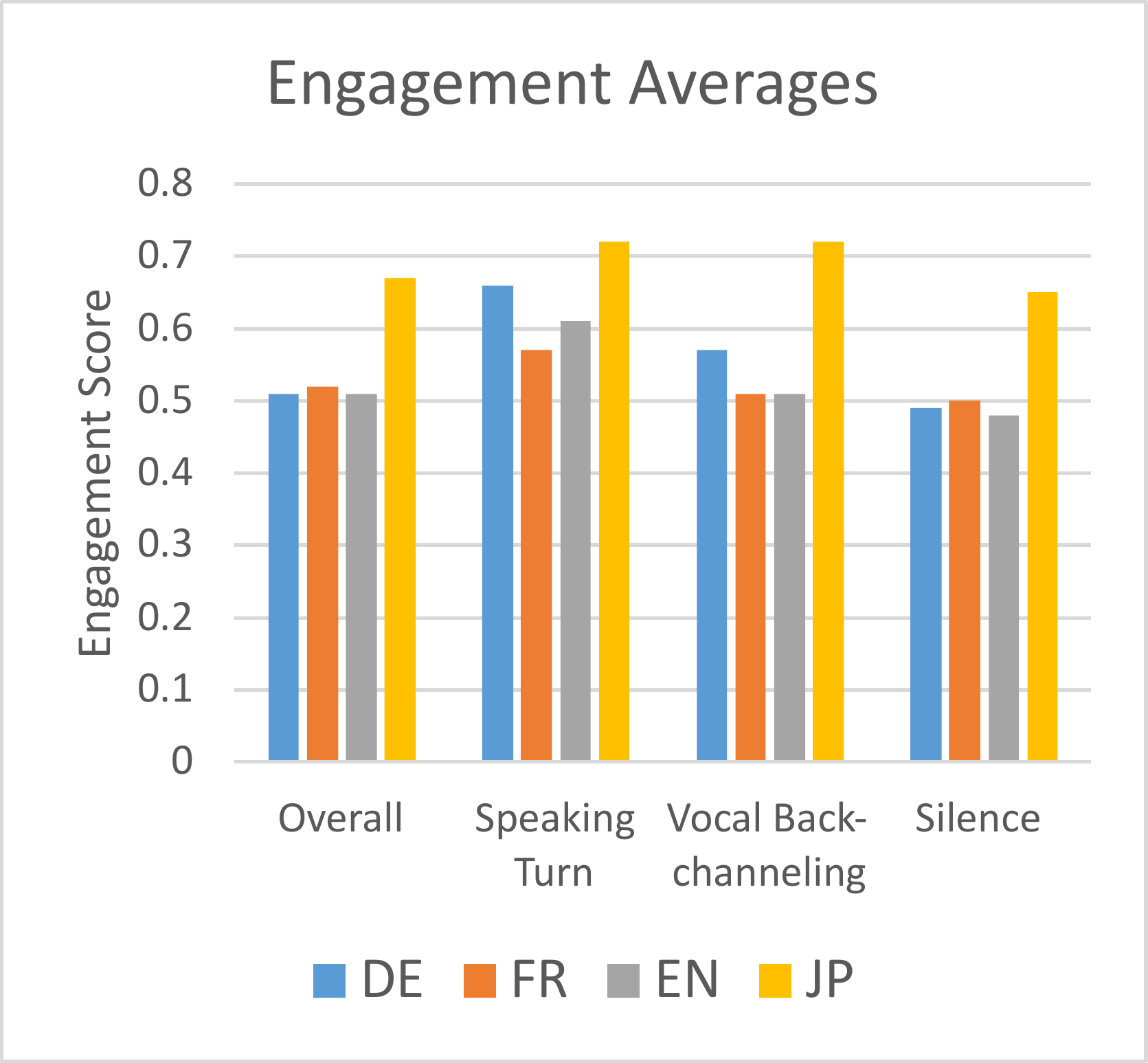}
    \caption{Novice engagement annotation averages overall and by speaking state.}
    \label{engavg}
  \end{minipage}
\end{figure}

Finally, we observed differences in the correlations between expert and novice engagement. In all European language sessions, the engagement of the novice and the engagement of the expert have a negative correlation coefficient (DE: r=$-0.19$, FR: r=$-0.05$, EN: r=$-0.07$). This suggests a slight tendency for one interlocutor's engagement to increase as the other's decreases and vice versa, although the weak correlations indicate that this mutual influence is minimal (see Figure \ref{engcor}). The minor adversarial effect may be explained by the difference in the average engagement of novices between when they hold the speaking turn compared to when they are silent, with a difference of more than 0.15 points on average for the German and English novices (see Figure \ref{engavg}) . 
The Japanese session, however, shows a high positive correlation coefficient (r=$0.51$) for mutual engagement. The data also show only a slight decline in engagement during silent intervals and no difference when vocally backchanneling in comparison to having the speaking turn.

\section{Engagement Prediction Experiment}

\begin{table*}[t]
\caption{Results of the initial trained models applied to every dataset. Engagement is abbreviated as \textit{E}, Mean Squared Error loss as \textit{MSE}. The marked models were trained on the same language as the test set. LSG describes a \textit{Language Speaker Group}.}
\resizebox{\textwidth}{!}{
\begin{tabular}{|c|c|c|c|c|c|c|}
\hline
\textbf{Model training set}&\textbf{MSE German LSG}&\textbf{MSE French LSG}&\textbf{MSE English LSG}&\textbf{MSE Japanese LSG} &\textbf{MSE European LSG}&\textbf{MSE Global}\\
\hline
German LSG& \textbf{0.008} & 0.020 & 0.024 &0.045&0.016& 0.023\\
French  LSG& 0.009 &\textbf{0.014} & 0.021 &0.027&0.013& 0.017 \\
English  LSG& 0.009 & 0.019 &\textbf{0.020}  &0.026&0.016& 0.018\\
Japanese LSG& 0.044& 0.046 & 0.046&\textbf{0.017}  &0.047& 0.040\\
\hline
European LSG& 0.008 & 0.015 & 0.020 & 0.033 & \textbf{0.013} & 0.018 \\
\hline
Global   & 0.011 & 0.019 & 0.022 & 0.014& 0.016& \textbf{0.016} \\
\hline

\end{tabular}
\label{resultTable}
}
\end{table*}

\begin{table*}[t]
\caption{Results after transfer learning to every cross reference dataset. Engagement is abbreviated as \textit{E}, Mean Squared Error loss as \textit{MSE}. The best results for each test set are marked. LSG describes a \textit{Language Speaker Group}.}
\resizebox{\textwidth}{!}{
\begin{tabular}{|c|c|c|c|c|c|c|c|}
\hline
\textbf{Initial model training set}&\textbf{MSE German LSG}&\textbf{MSE French LSG}&\textbf{MSE English LSG}&\textbf{MSE Japanese LSG} &\textbf{MSE European LSG}&\textbf{MSE Global LSG}\\
\hline
German LSG& \textbf{-} & 0.015 & 0.020 & 0.015 & 0.013 & 0.016\\
French LSG& 0.009 &\textbf{-}  & 0.020 & 0.016 & 0.014 & 0.016\\
English LSG& \textbf{0.008} & \textbf{0.014} &\textbf{-}  & \textbf{0.015} & \textbf{0.013} & \textbf{0.016}\\
Japanese LSG& 0.011 & 0.016 & 0.021 &\textbf{-}  & 0.014 & 0.019\\
\hline
European LSG& 0.008 & 0.015 & \textbf{0.019} & 0.015 & \textbf{-} & \textbf{0.016}\\
\hline
Global   & 0.010 & 0.018 & 0.022 & 0.017 & 0.015 & \textbf{-} \\
\hline

\end{tabular}
\label{transferTable}
}
\end{table*}

In this section, we examine the effect of the correlation between engagement and non-verbal behavior (see Figure \ref{engfeatcorr} and \ref{engavg}) on the accuracy of machine learning models in predicting engagement across different language speaker groups, hereafter called \textbf{LSG}s. 
Initially, we investigate how the differences in non-verbal behavior among different LSGs influence the accuracy of engagement prediction in a cross-corpus scenario (Section \ref{crossexp}). 
Subsequently, we assess the transfer-ability of model knowledge (parameters) related to non-verbal behavior across different LSGs using transfer learning. We discuss the adaptability of the non-verbal behavior-based estimation model across different LSGs by examining the improvements in accuracy facilitated by the transfer learning methodology (Section \ref{transferlearn}).

Finally, we analyze feature importance on the engagement prediction with SHAP values, showing clear consistency between the results of the data analysis and the importance the models assigns to their input features (Section \ref{Shap}).

\subsection{Features}

The prediction models were trained using 49 feature streams: 17 for facial action units, 20 for body properties, 3 for head angle movements, 3 for additional properties, expert engagement, and the computed features head nods, silence, vocal back-channeling, and speaking turn. 

The models were trained with engagement annotations by frame as the target value. To minimize annotator bias, we used the average of three annotators for all sessions except for 37 Japanese sessions. While these sessions are used for training the Japanese based prediction model, we do not use them for transfer learning.

\subsection{ML model and training procedure} 

We designed 4-celled LSTM models using a 30-frame window, each containing values for the 49 features to predict the engagement of the following frame. The frame window represents a temporal dimension that can capture changes such as head moments. We trained a regressive model and used the MSE to highlight performance differences between models and cultures. The data were split into training and testing set, with the first four sessions of each LSG serving as the testing set. Cross-culture predictions were also tested on the first four sessions of each specified LSG.

The dropout rate was set to 0.3 to prevent overfitting. The Data were randomized before training. For the initial training, the models were each trained for 5 epochs, with a learning rate of $4*10^{-4}$ . 
For the transfer learning to another LSG, we fine-tuned the models by adjusting the training length to 1 epoch and reducing the learning rate to $10^{-4}$.
We determined hyperparameters such as the learning rate and the number of epochs via grid search.

\subsection{Results}

\subsubsection{Cross-language corpus experiments} \label{crossexp}
The models were initially trained only for a single LSG. Most models perform best for the test set of the LSG they were trained on (see Table \ref{resultTable}). The worst-performing model is the English LSG model with a loss of 0.020, which performed only slightly better on the English test set than the French LSG model with a loss of 0.021. This is not surprising, as the English LSG training set is by far the most culturally diverse, with people from over 10 cultural backgrounds participating in the recordings, whereas German and French LSG recordings only have 3 and 4 cultural backgrounds respectively. This cultural diversity most likely also contributed to the English LSG model performing best on the Japanese LSG test set out of all European LSG models with a loss of 0.026. The Japanese LSG model performed poorly on all the other models, with a loss of 0.017 for the Japanese LSG test set, and a loss of over 0.040 for all other test sets.

We trained a model containing all the languages with 16 test sessions, named \textbf{Global} model, to revise the necessity of single LSG models. We also trained a model with training data from German,English and French LSG sessions and 12 test sessions called \textbf{European LSG} model, as the model performance, in addition to the ANOVA results, highlights a clear distinction between the European LSG and Japanese LSG parts of the dataset. While the European LSG model proved very adept in predicting engagement for all the European LSG sessions, even performing equally to the English LSG model on the English LSG test set with a loss of 0.020, it performed worse on the Japanese LSG test set than the French and English LSG models did. The Global model is more accurate than any other model on the Japanese LSG test set with a loss of 0.014, but is less capable of predicting engagement for the German, French and English LSGs than the European LSG model is.

\subsubsection{Transfer learning results} \label{transferlearn}

We then transfer learned each model to all other models and tested them on the respective training set.
The results (see Table \ref{transferTable}) show only minor improvements for inner European LSG model transfer learning. The German LSG model improved the most after the transfer learning on the French LSG training set, from a loss of 0.020 to a loss of 0.015 on the French test set.

Overall, transfer learning was most successful on the Japanese LSG training set, reducing the loss from 0.040–0.047 to 0.015–0.017.
Transfer learning on the Japanese LSG training set also substantially improved the performance of all other models on the Japanese LSG test set, sometimes outperforming the original Japanese LSG model. 
Surprisingly, the Global model's performance declined after transfer learning, with the loss increasing from 0.014 before transfer learning to 0.017 afterward.

\subsubsection{Feature analysis with SHAP values} \label{Shap}

Finally, we aimed to investigate which features influenced the models' decision-making processes. We used the SHAP (SHapley Additive exPlanations) method\footnote{https://github.com/shap/shap} to extract the SHAP values, which quantify the weight a model gives to each input feature, and compared them across the models (see Table \ref{SHAPTable}). 

First, we noticed that every model used fluidity as its primary factor for engagement prediction. This is surprising, given that fluidity showed a significant lack of correlation with engagement in the initial analysis  (DE: r=$0.02$, p<$0.001$; FR: r=$-0.02$, p<$0.001$; JP: r=$0.01$, p<$0.001$; EN: r=$0.00$, p=$0.01$). 

\begin{table}[t]
    \caption{Results of the SHAP analysis for all features for which a model showed a weight of 0.01 or higher.}
    \centering
    \resizebox{\columnwidth}{!}{
    \begin{tabular}{|l|c|c|c|c|c|c|}
    \hline
        &German LSG& French LSG& English LSG& Japanese LSG& Europe LSG& Global \\ \hline
        Fluidity & \textbf{0.062} & \textbf{0.055} & \textbf{0.024} & \textbf{0.097}& \textbf{0.111} & \textbf{0.044} \\ \hline
        Active speaking turn & \textbf{0.013} & 0.009 & 0.009 & 0.003 & \textbf{0.011} & \textbf{0.010} \\ \hline
        Silence & \textbf{0.011} & 0.009 & 0.008 & 0.005 & \textbf{0.011} & 0.008 \\ \hline
        Spatial extent & 0.009 & 0.007 & \textbf{0.010} & \textbf{0.015} & 0.004 & \textbf{0.010} \\ \hline
        Head yaw & 0.009 & 0.009 & 0.002 & \textbf{0.011} & 0.008 & 0.006 \\ \hline
        Head roll & 0.008 & 0.008 & 0.007 & \textbf{0.026} & 0.007 & 0.001 \\ \hline
        Overall activation & 0.007 & 0.007 & 0.009 & \textbf{0.026} & 0.01 & 0.002 \\ \hline
        Energy hands & 0.007 & 0.009 & 0.006 & \textbf{0.011} & 0.006 & 0.002 \\ \hline
        Expert Engagement & 0 & 0 & 0.002 & \textbf{0.010} & 0.002 & 0.007 \\ \hline
    \end{tabular}
    }
    \label{SHAPTable}
\end{table}

Second, the models with the European language speakers rely on \textit{Speaking Turn} and \textit{Silence} information (Section \ref{compfeat}) as relevant features with weights of between 0.008 and 0.013, whereas the model trained on the Japanese LSG attributes only had a marginal weight of 0.003 and 0.005 for those features.

The model trained on Japanese speakers considers head movement, energy of the hands, overall activation, and spatial extent as relevant features, which are also influential on all the European models to a lesser degree. The most prominent input feature of the Japanese LSG model is expert engagement, with a weight of 0.010, which the other LSG models consider irrelevant.
\section{Discussion}

We first noticed substantial differences between the European and Asian language sessions in the results of 
the Tukey-Kramer test. These results are in line with general findings such as Hall \cite{hall1976beyond}.

We found substantial differences in smiling frequency, especially a lower frequency in Chinese participants, similar to observations by Talhelm et al. \cite{TalhelmSmile} and Lu et al. \cite{LuSmile}. Additionally, there was a slightly greater frequency of smiles in the Japanese recordings that was correlated with engagement, which is not supported by literature, as many researchers deny a high correlation between smiles and the engagement of Japanese people \cite{Matsumoto1993}. 

The importance of factors such as overall activation, energy of the hands, the fluidity of movement, and the expression of emotion were already recognized by Wallbott \cite{BodilyexpressionWallbott} and found to have a substantial impact on engagement prediction. The lack of significant correlation between the fluidity of movement and engagement in the general analysis of the data might be ascribed to the model being able to recognize patterns over its 30 frame window that were missed in a frame-wise comparison.   

The computed head nods were not a relevant factor in engagement prediction. However, head movements in general were a relevant factor for all engagement models and target LSGs, and were most relevant for the Japanese language sessions. This is reflected in the extracted SHAP value attributed to head movement of the Japanese LSG model. While these findings are unambiguous, they do not completely reflect the literature, which suggests a strong disparity in backchanneling behavior and especially head nods in the Japanese data in comparison to the European data as described in Chapter \ref{Backchanneling}. 

Turn-taking has been found to have a substantial influence on engagement \cite{HsiaoEngagementTurn-taking}.  While we noticed a considerable difference in annotated engagement for the European sessions in the average of engagement for each speaking state, there were far less pronounced in the Japanese conversation annotations. This constitutes a considerable finding for the difference in engagement between cultures.

Finally, we observed a significant positive correlation between novice engagement and expert engagement for the Japanese recordings which was not present in the original NoXi sessions. This suggests a stronger need for harmony among the Japanese participants, leading them to conform more closely to the mood of their interlocutors. This findings aligns with Hofstede's theory of cultural dimensions \cite{Hofstede2001}, which attributes a higher degree of conformity to Japan than to Germany, France, or the UK.

We have found that the statistical findings of cultural differences in features are mostly reflected in the engagement prediction models, their accuracy and the improvement of model results after transfer learning.

\section{Conclusion}

We introduced \textbf{Noxi-J}, a new addition to the publicly available multi-lingual dyadic interaction corpus NoXi, which consists of a multimodal dataset featuring Japanese and Chinese speakers. Furthermore, we investigated the cultural variations in non-verbal features and their impact on engagement across different language groups and conducted comparative analyses. Finally, we trained an LSTM model for engagement prediction to verify the insights of the data analysis.

We focused on computed and automatically extracted features. Although the inclusion of manually annotated features might have helped identify further culture-specific variations, the high costs made this impractical for every feature of interest. Additionally, inner-group differences, especially within the dataset of English speaking participants, highlight the potential benefits of segregating the data on the basis of the participants' home culture.

The need for engaging and connecting with artificial systems is growing \cite{HowCulture}. Research has revealed issues in communication between different cultures caused by non-verbal communication \cite{BackchannelResponses}. Comprehensive data based analyses of cultural differences in non-verbal communication and backchanneling, as conducted here, are essential for the development of culturally sensitive agents and systems. 
This paper serves as an introduction, providing a baseline for more optimized engagement prediction models and acting as a reference point for further research into cultural differences in AI agents.

\section{Acknowledgments}

The research presented in the paper was conducted in the trilateral PANORAMA project and was partially funded by Deutsche Forschungsgemeinschaft (DFG), Project No. 442607480, and JST AIP Trilateral AI Research, Japan, Project No. JPMJCR20G6.

\bibliographystyle{ACM-Reference-Format}
\bibliography{ICMI_paper}
\typeout{get arXiv to do 4 passes: Label(s) may have changed. Rerun}

\end{document}